\documentclass[letterpaper, 10 pt, conference]{ieeeconf}  

\IEEEoverridecommandlockouts                              
                                                    
\usepackage{amsmath} 
\usepackage{amssymb}  
\usepackage{subcaption}

\usepackage{graphicx}
\graphicspath{ {./figures/} }
\usepackage[textsize=tiny]{todonotes}
\usepackage{url}

\usepackage[breaklinks]{hyperref}

\DeclareMathOperator*{\argmax}{arg\,max}
\DeclareMathOperator*{\argmin}{arg\,min}

\title{\LARGE \bf
CAPRICORN: Communication Aware Place Recognition using Interpretable Constellations of Objects in Robot Networks}

\author{Benjamin Ramtoula$^{1,2}$, Ricardo de Azambuja$^{1}$ and Giovanni Beltrame$^{1}$
\thanks{$^{1}$Department of Computer and Software Engineering, \'Ecole Polytechnique de Montr\'eal, Canada}%
\thanks{$^{2}$School of Engineering, \'Ecole Polytechnique F\'ed\'erale de Lausanne, Switzerland    }%
\thanks{{\tt\small contact: benjamin.ramtoula@polymtl.ca}}
}

\usepackage{cite}

\renewcommand{\theenumi}{(\alph{enumi}}

\begin{document}

\maketitle
\thispagestyle{empty}
\pagestyle{empty}

\begin{abstract}
  Using multiple robots for exploring and mapping environments can provide
  improved robustness and performance, but it can be difficult to
  implement. In particular, limited communication bandwidth is a considerable
  constraint when a robot needs to determine if it has visited a location that
  was previously explored by another robot, as it requires for robots to share descriptions of places they have visited.
  One way to compress this description is to use \emph{constellations}, groups
  of 3D points that correspond to the estimate of a set of relative object
  positions. Constellations maintain the same pattern from different
  viewpoints and can be robust to illumination changes or dynamic elements.
  We present a method to extract from these constellations 
  compact spatial and semantic descriptors of
  the objects in a scene. We use this representation in a 2-step
  decentralized loop closure verification: first, we distribute the compact semantic
  descriptors to determine which other robots might have seen scenes with
  similar objects; then we query matching robots with the full
  constellation to validate the match using geometric information.  The
  proposed method requires less memory, is more interpretable than global
  image descriptors, and could be useful for other tasks and
  interactions with the environment.  We validate our
  system's performance on a TUM RGB-D SLAM sequence and show its benefits in
  terms of bandwidth requirements. 
\end{abstract}
\section{INTRODUCTION}
\label{intro}

Simultaneous Localization and Mapping (SLAM) has applications in a variety
of scenarios: from aerial robots in GPS-denied environments to vacuum
cleaners and self-driving cars, supporting many different types of
missions. In some of these missions, a multi-robot SLAM system can
improve performance, robustness to individual robot failures, and allow the
use of heterogeneous robots with different means of movements and
sensors~\cite{saeedi_multiple-robot_2016}.

Multi-robot SLAM generally follows the same process as single-robot SLAM: each
robot uses sensor measurements to perform odometry and has a place
recognition system to close loops and correct odometry drift.  The
robots estimate global trajectories and generate maps solving an optimization
problem matching measurements and loop closures.

Robots perform odometry (the first step) locally and they are not affected by
any other robot collaborating for the SLAM task. Odometry can, therefore, follow
a method from the mature field of single-robot
SLAM~\cite{scaramuzza_visual_2011}.  However, during place recognition and
optimization, the robots need to consider information gathered from other robots
working in parallel to take advantage of the multi-robot system and build a
global map. For these steps, major adaptations of single robot SLAM techniques
are necessary to cope with challenges and constraints inherent to multi-robot
systems~\cite{saeedi_multiple-robot_2016}.

There are two extremes when considering sharing information among robots: a
centralized system, where a single entity communicates with every robot
performing operations as well as giving
feedback~\cite{forster_collaborative_2013,riazuelo_c2tam_2014,chong_moarslam_2016};
and a decentralized system, in which robots communicate directly.  Centralized
systems rely on the central entity to always be functional and scale in
computation and bandwidth with the number of robots.  On the other hand,
decentralized systems do not rely on a single entity and avoid the bottlenecks
of centralized systems~\cite{cieslewski_data-efficient_2018}.

\begin{figure}[tpb]
\centering
\includegraphics[width=0.42\textwidth]{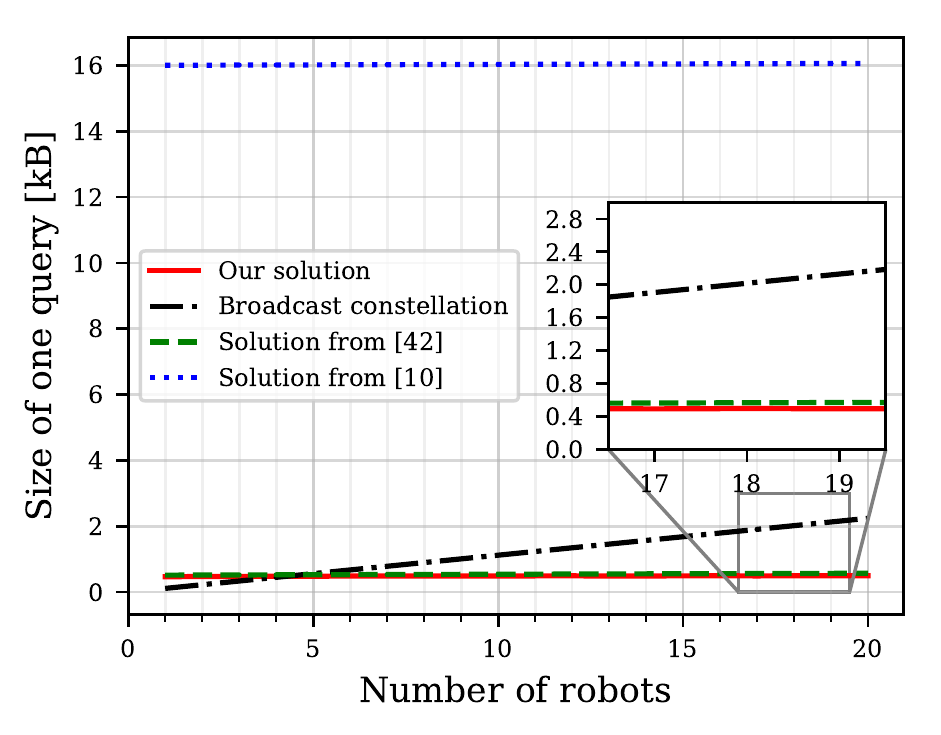}
\caption{Size of a place recognition query over the full team depending on number of robots. ``Broadcast constellation'' corresponds to sending the full constellation to all robots. 1-hop communication between all robots is considered here.}
\label{fig:data_exchange}
\end{figure}

\begin{figure*}[t!] 
  \centering
  \includegraphics[width=17cm,height=5cm]{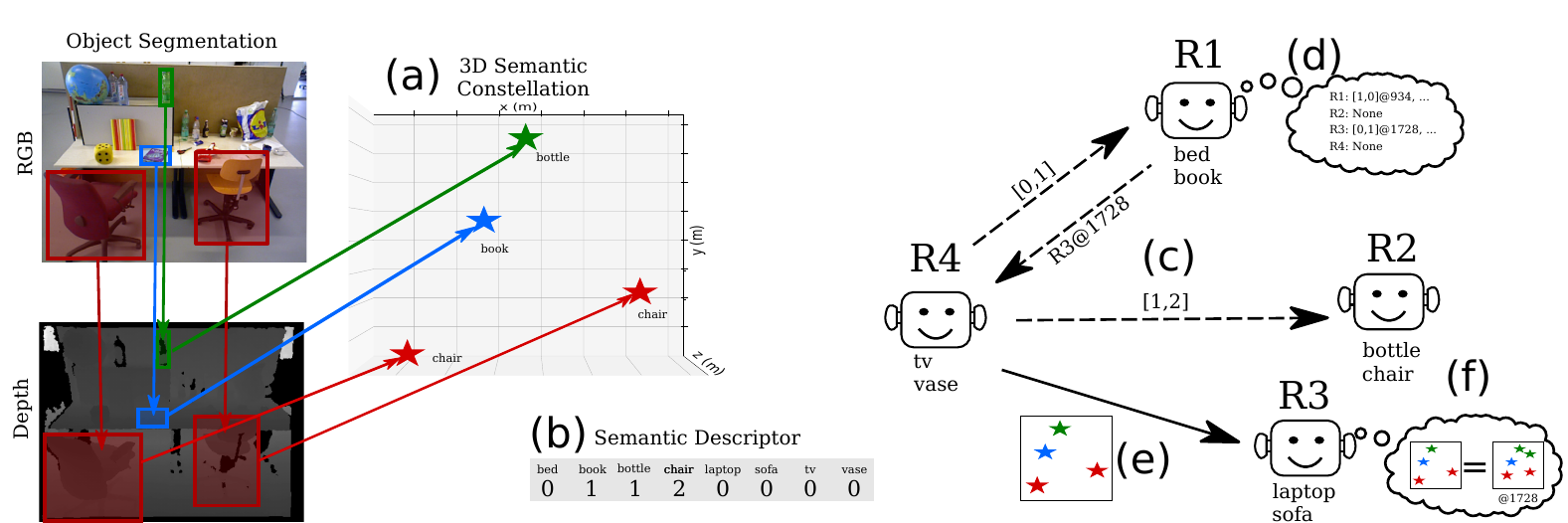}
  \caption{Illustration of our full system, in a simplified case with 4 robots. In this example, the detector can predict 8 classes, and R4 is making a query. (a) R4 creates a constellation. (b) R4 generates an associated semantic descriptor. (c) R4 splits and distributes the descriptor among robots responsible for classes seen: R1 (book), and R2 (bottle and chair). (d) R1 and R2 save the sub-descriptors, compare them against all previously received sub-descriptors, and respond with match candidates: R1 responds that R3 at frame 1728 had sent a similar sub-descriptor. (e) R4 sends its full constellation to R3 since it matches best semantically. (f) R3 performs data association between constellations using our geometric surroundings descriptors and produces a final loop closure score. Details of the process including descriptors and comparisons are presented in Section~\ref{sec:methods}.}
  \label{fig:overview}
\end{figure*}

In this paper, we present a method for Communication Aware Place Recognition using Interpretable Constellations of Objects in Robot Networks (CAPRICORN).
We introduce compact spatial and semantic descriptors which can be compared for place recognition using constellations, a representation of the environment which is particularly well suited for the constraints of multi-robot systems. We also present a mechanism to decentralize it efficiently (Fig.~\ref{fig:data_exchange}).

Based on previous ideas~\cite{finman_physical_2014,frey_efficient_2018}, and
considering the recent advances in real-time object detection methods, RGB-D
sensors and embedded AI systems, we build 3D \emph{constellations} of points
to represent scenes, where each point approximates the position of an
object. The constellation is human understandable and capable of supporting 
other tasks requiring interaction with the environment.
Its shape and scale are robust to viewpoint changes.
Since a constellation builds on an object
detector module, it can inherit robustness properties which can be associated
with the detector, such as invariance to illumination changes. Our method
benefits from the fact that object representations require significantly less memory and bandwidth than other representations
based on visual features or point clouds~\cite{choudhary_multi_2017}.

Constellations allow us to perform a decentralized loop closure check in 2
steps. To find loop closures, we first use a semantic descriptor of a
constellation which is extremely compact and contains only information
concerning object labels. Inspired by the method
in~\cite{cieslewski_efficient_2017-bow}, each robot shares the descriptor with all
other robots, each of which potentially finds similar scenes based on specific
object labels, and returns candidate scene matches. From these results, the
querying robot sends its constellation to the potential matching
robots. These, in turn, compute an overall score considering both ``which'' semantic
elements are two frames (labels), and ``how'' they are organized in space
(relative distances), allowing accurate loop closure detection. The process is
illustrated in Fig.~\ref{fig:overview} and detailed in
Section~\ref{sec:methods}, while experiments and final results are presented
in Sections~\ref{sec:exper} and~\ref{sec:results},
respectively. The source code is provided on our repository: \url{https://github.com/MISTLab/CAPRICORN}, and a brief summary of this work is presented in~\cite{ramtoula-mrs_capricorn}.

\section{RELATED WORK}
\label{rel_work}

\subsection{Visual Place Recognition}
A major part of a visual place recognition module is the description of a
place~\cite{lowry_visual_2016}. Techniques using solely visual features,
either global~\cite{oliva_chapter_2006,ulrich_appearance-based_2000} or
local~\cite{lowe_object_1999,leonardis_surf_2006}, that can be quantized with
bag-of-words~\cite{fei-fei_li_bayesian_2005,sivic_video_2003} are most
commonly used. However, an important challenge is for the descriptors to
exhibit robustness to extreme viewpoint changes as well as dynamic changes in
the environment. These changes can severely affect the visual appearance of an
image, making methods relying only on visual features fail.  Advances in deep
learning, specifically convolutional neural networks
(CNNs)~\cite{krizhevsky_imagenet_2012}, allow to extract more robust and
high-level features from
images~\cite{redmon_you_2016,long_fully_2015,ren_faster_2017,liu_ssd_2016}.
Recent works propose various methods to make use of CNNs to improve the place
recognition performance. Some of
these~\cite{chen_only_2017,garg_dont_2018,garg_lost_2018,suenderhauf_place_2015,naseer_semantics-aware_2017}
make use of learned feature maps of CNNs trained for other tasks such as object proposals, which can be robust to viewpoint and
condition
changes~\cite{suenderhauf_performance_2015}. NetVLAD~\cite{arandjelovic_netvlad_2016}
is a CNN architecture aimed at creating a global descriptor based on an
image, trained end-to-end for place recognition.
While these methods improve robustness, they lack interpretability and do not
consider computation and communication constraints, which can be significant
in a multi-robot system.

\subsection{Semantic entities for mapping and place recognition}
Using semantic entities in the environment
for mapping has been gaining popularity in recent years.
SLAM++ is an early work which proposed a full SLAM solution based on a set
of predefined objects whose 3D models are
available~\cite{salas-moreno_slam++_2013}.

Following works avoided the need for object models by using discovery from point cloud data~\cite{choudhary_slam_2014}, 2D object detections to create 3D models on the go~\cite{mccormac_fusion++_2018}, or by fitting 3D quadrics to
objects~\cite{nicholson_quadricslam_2019,hosseinzadeh_structure_2018,hosseinzadeh_real-time_2018}. These
SLAM solutions require precise estimates of the objects' poses for loop closures.

Several solutions consider loop closures using graphs or constellations of
entities.  Finman et
al. detect objects using primitive kernels to build a
graph~\cite{finman_efficient_2014}, from which they perform place recognition.  Gawel et al.  propose to build a graph from
semantically segmented images for multi-view
localization~\cite{gawel_x-view_2018}. Frey et al. consider constellations of
landmarks which can be matched geometrically to perform data association and map
merging in semantic SLAM~\cite{frey_efficient_2018}.

\subsection{Decentralized place recognition}
The problem of decentralized place recognition requires a mechanism to share
descriptions efficiently. When full connectivity cannot be assumed, some methods rely on sharing information only to
neighboring
robots~\cite{cunningham_ddf-sam_2013,montijano_distributed_2013,franchi_sensor-based_2009},
to deduce global associations.
For fully connected teams, Cieslewski and Scaramuzza
proposed a way to share a bag-of-words descriptor that is split among every
robot such that the amount of data exchanged for a place recognition query is
independent of the number of robots~\cite{cieslewski_efficient_2017-bow}. They
extended their work using NetVLAD \cite{arandjelovic_netvlad_2016} as a
global descriptor, with a mechanism allowing each descriptor to be sent only
to one specific robot~\cite{cieslewski_efficient_2017-netvlad}. These
solutions usually make use of descriptors which are hardly interpretable and not reusable for other tasks.

\section{METHODOLOGY}
\label{sec:methods}

Our decentralized place recognition method (Fig.~\ref{fig:overview}) relies on the creation of a spatially and
semantically meaningful descriptor (a \emph{3D constellation of objects})
associated with each frame coming from an on-board RGB-D camera.   
Similarly to the work that inspired us using visual
descriptors\cite{cieslewski_efficient_2017-bow}, the sharing process requires
to pre-assign all possible classes of detectable objects among the
robots in the system.

Overall, for each RGB-D frame, we execute the following steps, as illustrated in Fig.~\ref{fig:overview}:
\begin{enumerate}
    \item A querying robot QR creates a constellation~(\ref{subsec:constellation})
    \item QR generates an associated semantic descriptor~(\ref{subsec:semantic_desc_creation})
    \item QR distributes the descriptors among a subset of the other robots,
      based on the objects in the scene~(\ref{subsec:semantic_desc_share})
    \item The receiving robots save the descriptors, compute their scores against
      all previously received descriptors, and respond with match candidates~(\ref{subsec:semantic_desc_share})
    \item QR sends its full constellation to the robots with the
      closest-matching global semantic
      descriptors~(\ref{subsec:semantic_glob_check})
     \item The receiving robots selected in step (e) perform data association between the
       constellations~(\ref{subsec:data_assoc}) and combine the results with
       global semantic descriptors to make a place recognition
       decision~(\ref{subsec:full_check})
\end{enumerate}

\subsection{3D Constellations of Objects}
\label{subsec:constellation}
Our system consists of a robot set
$\Omega = \{\alpha,\beta,\gamma \ldots \}$, where each robot $\omega$ has
registered a set of frames $I_\omega$. Each frame $i$ contains 3 color channels (RGB) and one depth channel: $i \doteq \{i_R,i_G,i_B,i_D\}$.

From each frame $i \in \{I_\alpha, I_\beta, I_\gamma, \ldots \}$, an object-to-point detection system detects visible objects and associates them to a 3D point expressed in the camera frame. 
The implementation of this system is left as a design choice depending on types of environments, computing power, and algorithms available. 

Each point $k$
generated from a detected object in frame $i$ of robot $\alpha$ is associated
with a label $l_{\alpha_i}^k$ taken within $L$, the set of all detectable indexes which depends on the object-to-point semantic segmentation system. A point also has a 3D position $\mathbf{p}_{\alpha_i}^k$.
We define an object as
$o_{\alpha_i}^k\doteq \{l_{\alpha_i}^k, \mathbf{p}_{\alpha_i}^k
\}$. Therefore, for a frame $i$ seen by robot $\alpha$ we obtain the constellation
$\mathbf{\mathcal{C}}_{\alpha_i} \doteq \{o_{\alpha_i}^0, o_{\alpha_i}^1,
o_{\alpha_i}^2, \ldots \}$.

Since the object-to-point detection system is exchangeable, the proposed approach is straightforward to adapt to new environments with
different classes of objects, or simply to improve as
state-of-the-art detection tools advance.

\subsection{Semantic descriptor}
\label{subsec:semantic_desc_creation}
We design a semantic descriptor that does not use the 3D information of the
constellations to check for semantic consistency. It allows to compress information and to
simplify how information is split among robots.  This simplified descriptor
can be seen as a histogram in which each bin corresponds to a possible class
in $L$. A bin's value is the number of instances of the class in the scene. For constellation $\mathbf{\mathcal{C}}_{\alpha_i}$, the semantic descriptor $\mathbf{\mathcal{S}}_{\alpha_i}$ has a value in the bin corresponding to class index $l$ given by $\mathcal{S}_{\alpha_i}[l] = | \{o_{\alpha_i}^{k} | l_{\alpha_i}^k = l \;
\forall k \}\ \! | $ such that
$\mathcal{S}_{\alpha_i} \in \mathbb{N}^{|L|}$.  This semantic
descriptor $\mathcal{S}_{\alpha_i}$ is very sparse
since most classes which can be detected in standard object detectors usually
do not appear in the same scene. It differs from the similar semantic words used in SemanticSIFT~\cite{arandjelovic2014visual} in that it counts the number of independent entities of a class rather than use a binary indicator of the presence of pixels from a semantic class.

\subsection{Distributed semantics}
\label{subsec:semantic_desc_share}
To check for similar places seen by other robots in an efficient, scalable, and
decentralized way, we use a technique from previous
work~\cite{cieslewski_efficient_2017-bow} which we adapt to our descriptors. The idea is to split the global
descriptor into partial descriptors which only consider specific classes, and
assign them to different robots.

During a query of frame $i$ from robot $\alpha$, each robot $\omega$, with
preassigned label indexes $\l^\omega \subset L$, is responsible for the
partial descriptor
$\mathcal{S}_{\alpha_i}^\omega = \mathcal{S}_{\alpha_i}[\l^\omega]$.  Robot
$\omega$ only receives non-zero values of the partial descriptors to reduce
communication bandwidth, benefiting from the sparsity of the semantic
descriptor. The receiving robots store the received partial descriptor and compare it to all previously
received ones. The comparison of two partial descriptors of $\mathbf{\mathcal{C}}_{\alpha_i}$ and
$\mathbf{\mathcal{C}}_{\beta_j}$ assigned to robot $\omega$ is performed using: 
\begin{align}
    \label{eq:jaccard_partial}
    s^\omega_{\alpha_i,\beta_j} = \frac{\sum_{l \in \l^\omega}\min({\mathcal{S}_{\alpha_i}^\omega[l]},{\mathcal{S}_{\beta_j}^\omega[l]})}{\sum_{l \in \l^\omega}\max({\mathcal{S}_{\alpha_i}^\omega[l]},{\mathcal{S}_{\beta_j}^\omega[l]})}
\end{align}
where $s^\omega_{\alpha_i,\beta_j}$ corresponds to the well-known Jaccard index of the descriptors, also
known as Intersection over Union (IOU), commonly used in object detection
\cite{redmon_yolov3_2018}.

Each robot $\omega$ then finds frames producing the highest scores and returns the corresponding frame and robot ID to the querying robot $\alpha$:
$(\gamma,m) = \argmax_{\beta,j} s^\omega_{\alpha_i,\beta_j}$. The number of highest
scoring frames returned $n_{ret}$ can be adjusted to the available bandwidth and number
of robots.  Note that the querying robot is also assigned a subset of labels,
and has registered previous queries related to those labels. Hence, it 
also needs to compare these to its own current query, which requires no external communication.

\subsection{Deciding on the best global semantic matches}
\label{subsec:semantic_glob_check}
The querying robot $\alpha$ receives robot and frame IDs from all robots
detecting partial semantic matches. $\alpha$ must decide which robots to
further query for a semantic and geometric comparison of the constellations.
Similarly to the geometric check step used in previous works
\cite{cieslewski_efficient_2017-netvlad,cieslewski_efficient_2017-bow}, we
choose to send the full constellation to the robots which provide the highest
number of frames returned from the highest scoring partial semantic vectors,
as these are more likely to have observed similar scenes. The robots receiving
the full constellation compare it to their own recorded constellations.

Again, the number of robots that should be further queried $n_{fq}$ can be
controlled and adjusted to the available bandwidth and the number of robots.

\subsection{Data association}
\label{subsec:data_assoc}
Once a robot $\beta$ receives a querying constellation
$\mathbf{\mathcal{C}}_{\alpha_i}$ from another robot $\alpha$, it compares
it to every constellation it has previously recorded.  The first step when
comparing constellations $\mathbf{\mathcal{C}}_{\alpha_i}$ and
$\mathbf{\mathcal{C}}_{\beta_j}$ is to perform data association: the process
of matching points of one constellation to points in another. In other
terms, matching specific objects detected in one frame to objects detected in
another.

Labels appearing in both constellations are defined as
$l_{\alpha_i,\beta_j} \subset L$. For each object $k$ with labels in
$l_{\alpha_i,\beta_j}$, we define a geometric surroundings descriptor
$\mathbf{v_{\omega_m}^k} \in \mathbb{R}^{|L|}$. Elements in this vector are
given the value $0.0$ if they correspond to the index of a label not appearing
in both constellations. Otherwise, elements are given by the Euclidean distance to the
closest instance of the corresponding label in the constellation:
\begin{align}
\mathbf{v_{\alpha_i}^k}[l] = \begin{cases}
0.0 &\text{if }l \notin l_{\alpha_i,\beta_j}\\
\min_{n | l_{\alpha_i}^n = l} ||\mathbf{p}_{\alpha_i}^k - \mathbf{p}_{\alpha_i}^n|| &\text{if }l \in l_{\alpha_i,\beta_j}
\end{cases}
\forall l \in L
\end{align}

This vector represents how far the closest elements of each appearing label
are from a given object.  We use this vector to find the most likely matches
between $\mathbf{\mathcal{C}}_{\alpha_i}$ and
$\mathbf{\mathcal{C}}_{\beta_j}$, that is finding points with the same label
and the closest vectors:
\begin{align}
    \text{match}_{\alpha_i,\beta_j}^k = \argmin_{n | l_{\beta_j}^n  = l_{\alpha_i}^k } ||\mathbf{v_{\alpha_i}^k} - \mathbf{v_{\beta_j}^n}||
\end{align}

We then keep the match between the object of index $k$ in
$\mathbf{\mathcal{C}}_{\alpha_i}$ and the object of index $n$ of
$\mathbf{\mathcal{C}}_{\beta_j}$ only if they both match each other and their
vectors are sufficiently close according to a threshold $d$:
\begin{align}
    \text{matches}_{\alpha_i,\beta_j}^m = (k,n) \text{ if }
     \begin{cases}
        n = \text{match}_{\alpha_i,\beta_j}^k\\
        k = \text{match}_{\beta_j,\alpha_i}^n\\
        ||\mathbf{v_{\alpha_i}^k} - \mathbf{v_{\beta_j}^n}|| < d
     \end{cases}
\end{align}
The value of $d$ depends on the estimation noise when reducing an object to a
point. $d$ must be high enough so that the noise does not prevent matches, but
small enough to reject incorrect matches. It can be tuned with prior experiments by measuring the variance of the estimated point position associated with an object when the viewpoint changes.

Our proposed method combines semantic information and geometric information
while being independent of the reference frame from which the 3D positions of
points are expressed. It can detect valid matches and filter out incorrect
ones while not requiring additional descriptors associated with each element of
the constellations.

\subsection{Loop closure detection}
\label{subsec:full_check}

The matching process (Section~\ref{subsec:data_assoc}) acts as a geometric
filter on valid semantic matches: only items of the same class and with
similar geometries of surrounding objects are matched. We use this to evaluate
a score between two constellations.  To obtain a full comparison of two
constellations, we compute the Jaccard index over the full semantic
descriptors of both constellations (Eq.~\ref{eq:jaccard_partial}, but with
$s_{\alpha_i,\beta_j}$ calculated using $\mathcal{S}_{\alpha_i}$ and
$\mathcal{S}_{\beta_j}$). This provides a semantic comparison to which we
multiply $g_{\alpha_i,\beta_j}$, the fraction of objects successfully matched
over the number of common objects in the two constellations. Using this
measure, for a place to be recognized we need two constellations to
demonstrate similar semantic entities resulting in a high
$s_{\alpha_i,\beta_j}$, but also for those entities to have similar geometric
relationships to produce a high $g_{\alpha_i,\beta_j}$.
\begin{align}
s_{\alpha_i,\beta_j} = \frac{\sum_{l \in L}\min({\mathcal{S}_{\alpha_i}[l]},{\mathcal{S}_{\beta_j}[l]})}{\sum_{l \in L}\max({\mathcal{S}_{\alpha_i}[l]},{\mathcal{S}_{\beta_j}[l]})}\\
g_{\alpha_i,\beta_j} = \frac{|\text{matches}_{\alpha_i,\beta_j}|}{\sum_{l \in L}\min({\mathcal{S}_{\alpha_i}[l]},{\mathcal{S}_{\beta_j}[l]})}\\
\text{score}_{\alpha_i,\beta_j} = s_{\alpha_i,\beta_j}. g_{\alpha_i,\beta_j}
\end{align}

\section{EXPERIMENTS}
\label{sec:exper}
\subsection{Dataset used and ground-truth}
\label{sec:exper:gt}
To ensure that our method is able to perform place recognition while also being data-efficient, we test with it using the \emph{freiburg3\_long\_office\_household} sequence of the TUM RGB-D SLAM dataset \cite{sturm_benchmark_2012}. To the best of our knowledge, no other datasets provide RGB-D data of environments containing objects and loop closures. As a proof-of-concept, our experiments are based on one sequence since loop closures had to be manually annotated.

The \emph{freiburg3\_long\_office\_household} sequence consists of 2585~frames recorded with a handheld camera circling around two back-to-back desks with a variety of different objects on them. The alignment of the reference frame with the desks allowed us to combine ground-truth pose of the camera to manual labeling in order to obtain a ground-truth of recognized places. We use the pose of the camera and the depth values to estimate the position of the scene observed in each frame. We can then use these scene positions to estimate distances between scenes observed in each pair of frames, and we obtain a ground-truth loop closure score based on these distances. We transform the distances to obtain scores between $0.0$, when scenes are very far, and $1.0$, when the scenes' positions overlap. We verify qualitatively the ground-truth scores by visualizing scores from samples of image pairs.
As consecutive frames can easily be detected as loop closures, we ignore closures from frames within 12 seconds (200 frames).

\subsection{Choices specific to our experiment}
\subsubsection{Object-to-point detection system}
Given a frame $i$, the channels $ \{i_R,i_G,i_B\}$ are given to the real-time object detector YOLOv3~\cite{redmon_yolov3_2018} with a zero threshold to detect as many objects as possible. We use the weights supplied by the authors obtained from training on the COCO dataset~\cite{lin_microsoft_2014}, which can detect $80$ objects: $L = \{0,1,2,...,78,79 \}$. Using the depth channel $i_D$, we compute the median 3D positions of points within the detected bounding boxes of objects to estimate a point associated to each object.

\subsubsection{Preassigning labels}
In this work we use a straightforward approach to assign labels to different robots.	Each robot is responsible for consecutive label indexes so that all robots have the same number of labels. We leave the study of more complex assignment strategies for future works.

\subsubsection{Parameter choice}
Based on our experiments, we allow each robot to return up to the $n_{ret} = 4$ highest matching frames when performing the partial semantic comparison (Section~\ref{subsec:semantic_desc_share}), as well as sending its full constellation to up to $n_{fq} = 4$ robots for the geometric and semantic check (Section~\ref{subsec:full_check}). This presented a good balance between performance and communication. We used a value of $0.25$m for $d$ to approve a match while performing data association.

\subsection{Evaluation}
We simulate our method in a computer program processing all the frames of the
\emph{freiburg3\_long\_office\_household} sequence. The program simulates all
the communications which would appear between robots as well as the
computations happening on each robot. It outputs the final loop closure scores
for each pair of frames.  We perform the evaluation of our system in
centralized and decentralized cases to verify that our decentralized mechanism
does not compromise performance.

In the centralized system, every constellation obtained can be compared to every other constellation registered during the previous frames of the sequence, and only the full comparison of two constellations is performed (\ref{subsec:full_check}).

\begin{figure}[tpb]
\centering
\includegraphics[width=0.3\textwidth]{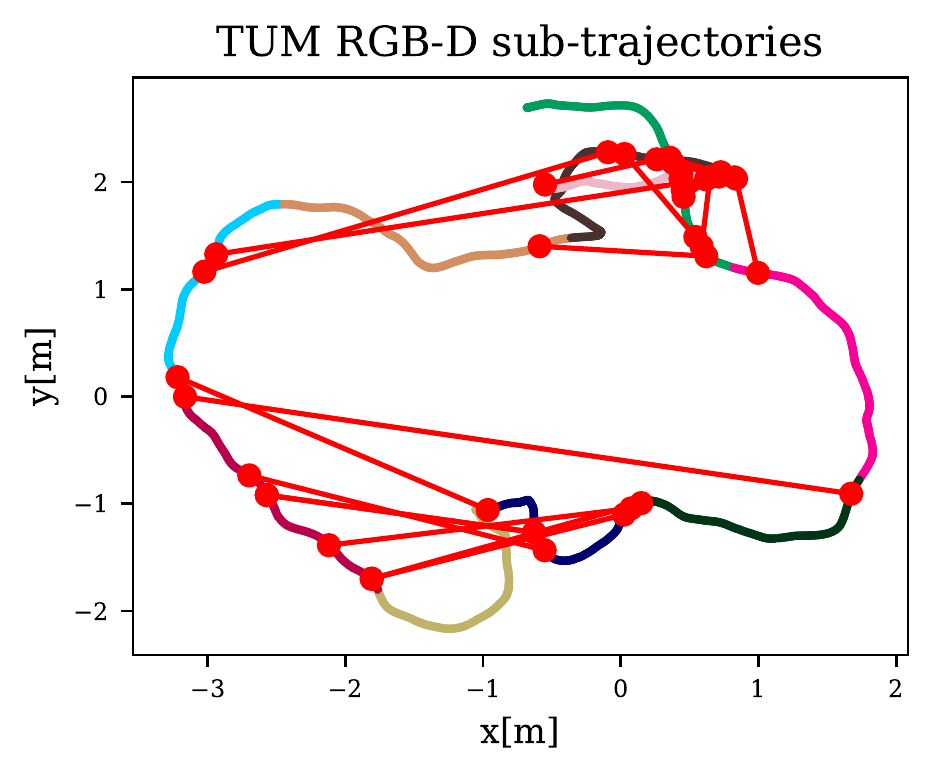}
\caption{Example of the \emph{freiburg3\_long\_office\_household} trajectory split between 10 robots. Red lines indicate detected loop closures when the camera was facing the same area.}
\label{fig:split_traj}
\end{figure}

For the decentralized case, we use a similar approach as \cite{cieslewski_efficient_2017-netvlad} by splitting the sequence into sub-trajectories, each assigned to a different robot. An example of a split sequence is shown in Fig.~\ref{fig:split_traj}. Moreover, instead of using the method on multiple processes, we simulate the data each robot would have access to and would communicate with. We evaluate both performance and data exchanged between robots. The data used by our representation is shown in Table~\ref{tab:data}.

\begin{table}[h]
\caption{Data required for our representations in bytes. Messages which are communicated are shown in bold font.}
\label{tab:data}
\begin{center}
\resizebox{0.45\textwidth}{!}{%
\begin{tabular}{|c||c|}
\hline
Value in the semantic descriptor                    & $0.5$                                   \\
\hline
Label index                                         & $1$                                     \\
\hline
\textbf{Distributed semantic query}                                & number of classes seen $\cdot (1+0.5)$               \\
\hline
Robot index                                         & $1$                                     \\
\hline
Frame index                                         & $2$                                     \\
\hline
\textbf{Returned robots and frame id} & $\leq n_{ret} \cdot(1+2)$                \\
\hline
Position value (x,y or z)                           & $2$                                     \\
\hline
\textbf{Full check query (constellation)}                      & $\leq  n_q \cdot$ number of objects $\cdot (1+3\cdot2)$ \\ 
\hline
\end{tabular}}
\end{center}
\end{table}

\section{RESULTS}
\label{sec:results}

\begin{figure}[tpb]
\centering
\includegraphics[width=0.49\textwidth]{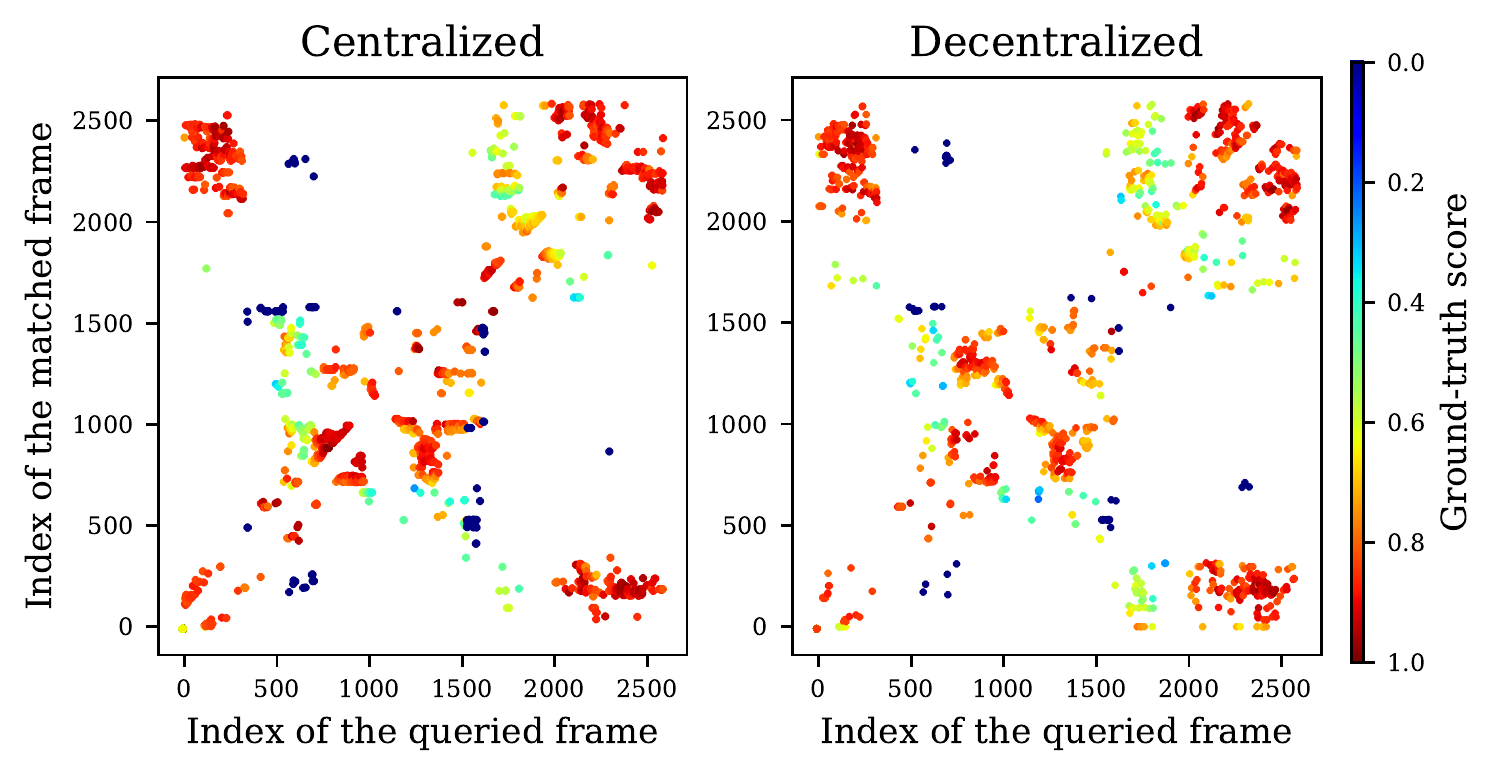}
\caption{Similarity matrices obtained for the centralized solution and decentralized solution with 10 robots. The diagonals are empty due to the removal of neighboring frames during the search. Points show the pairs producing the best loop closure score. Loop closure scores below $0.25$ were removed.
}
\label{fig:similarity_matrices}
\end{figure}

The similarity matrices obtained in both centralized and decentralized cases
are shown in Fig.~\ref{fig:similarity_matrices}. Overall, we can confirm that
similar regions of frame pairs produce detected loop closures in both
cases. The areas match what we expect from the sequence: the first desk is
seen at the beginning with various points of views until frame 450, and
then later after frame 1550. The second desk is visible, again with various
points of view, between frames 450 and 1550, which also corresponds to a region
with consistent matches. 
Most of the ground truth scores (Section~\ref{sec:exper:gt}) associated with the detected loop closures are close to $1.0$, and concentrated in particular regions, which confirms their validity. Yet, some smaller regions of detected closures are associated to a lower ground truth score. 
Even though a high ground truth score is a good confirmation of the match, a lower score doesn't necessarily mean that the matches are incorrect. A lower score means that estimated regions where the camera was facing in the two frames were considered to be far. However, we have no way of automatically and accurately checking if parts of two frames correspond to the same scene. It is possible that the camera was aimed at different areas which we have calculated to be relatively far, but that part of a same scene was visible. This could lead our system to recognize parts of constellations which would be visible in both frames, even though the ground truth score would be low. 

\begin{figure}[tpb]
\centering
\includegraphics[width=0.35\textwidth]{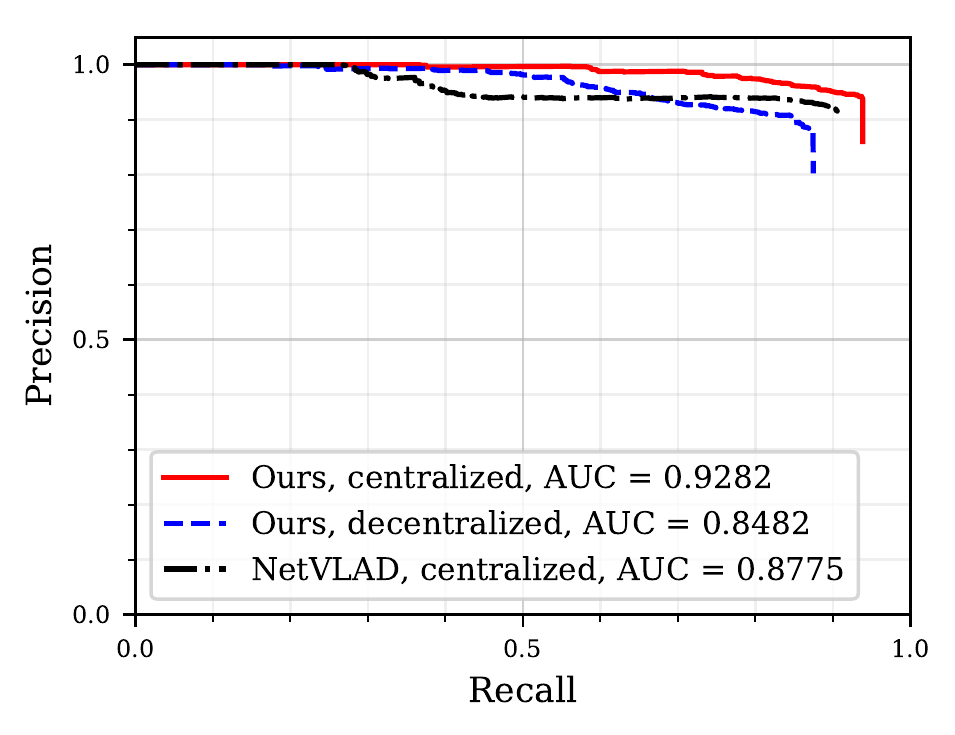}
\caption{Precision recall obtained in both centralized and
  decentralized cases (10 robots), compared with NetVLAD.}
\label{fig:precision_recall}
\end{figure}

The performance of our system is confirmed by looking at the precision-recall
curves, shown in Fig.~\ref{fig:precision_recall}. Our centralized system
performs very well, obtaining an area under curve (AUC) of $0.9282$. The
decentralized mechanism slightly affects the performance, reducing the AUC by
$8.6\%$. This drop is close to the drop of $6.7\%$ due to the decentralized
mechanism in the solution based on
NetVLAD~\cite{cieslewski_efficient_2017-netvlad}. Additionally, we evaluate a
centralized solution using the NetVLAD implementation from
DSLAM~\cite{cieslewski_data-efficient_2018}, which performs slightly better
than our decentralized solution but not as well as our centralized one.  The
best recall obtained from the decentralized solution is not as high as the one
from the centralized solution. This can be explained as a cost of the
decentralization mechanism. In fact, the mechanism will only perform full
checks on pairs of frames which were selected on a series of partial semantic
comparisons, whereas the centralized version will check every pair. As a
result, it is possible that some potentially good matches are ignored in the
decentralized case because they don't stand out using the series of partial
semantic comparisons. This situation can happen, for example, if two scenes contain common combinations of objects. The objects would be seen often and so the
two scenes won't produce particularly good partial semantic scores, and won't
be explored further. In the centralized case, those scenes are still checked completely
and produce high loop closure scores when considering data
association and the geometric score.

We also produce a comparison of data used for our decentralized method and
previous
works~\cite{cieslewski_efficient_2017-netvlad,cieslewski_efficient_2017-bow},
as well as the case where each robot broadcasts its constellation to every
other robot in Fig.~\ref{fig:data_exchange}. For previous works, the query size is independent of the sequence used, hence we can estimate the amount of data based on the mechanisms proposed. It is clear that all methods making use of a decentralization mechanism do not consume more data as robots are
added. Moreover, we
achieve query sizes of around 490B, which is comparable to previous work using
NetVLAD~\cite{cieslewski_efficient_2017-netvlad} requiring 512B. However, note that the close score between our solution and the one
based on NetVLAD is due to the fact that in the latter, the descriptor is sent
to only one robot. Considering cases in which robots may not be able to
communicate directly to every other robot, multi-hop communication may be
necessary. In that case, sending the NetVLAD descriptor of one frame using 512
bytes through several robots will quickly increase necessary bandwidth. In our solution, only very small messages are sent to every robot in the distributed semantic comparison and when performing the full comparison (Table~\ref{tab:data}). In the cases where multi-hop communication is required, our solution will scale better with the number of hops required.

\begin{figure}[tpb]
\centering
\includegraphics[height=0.2695256\textwidth]{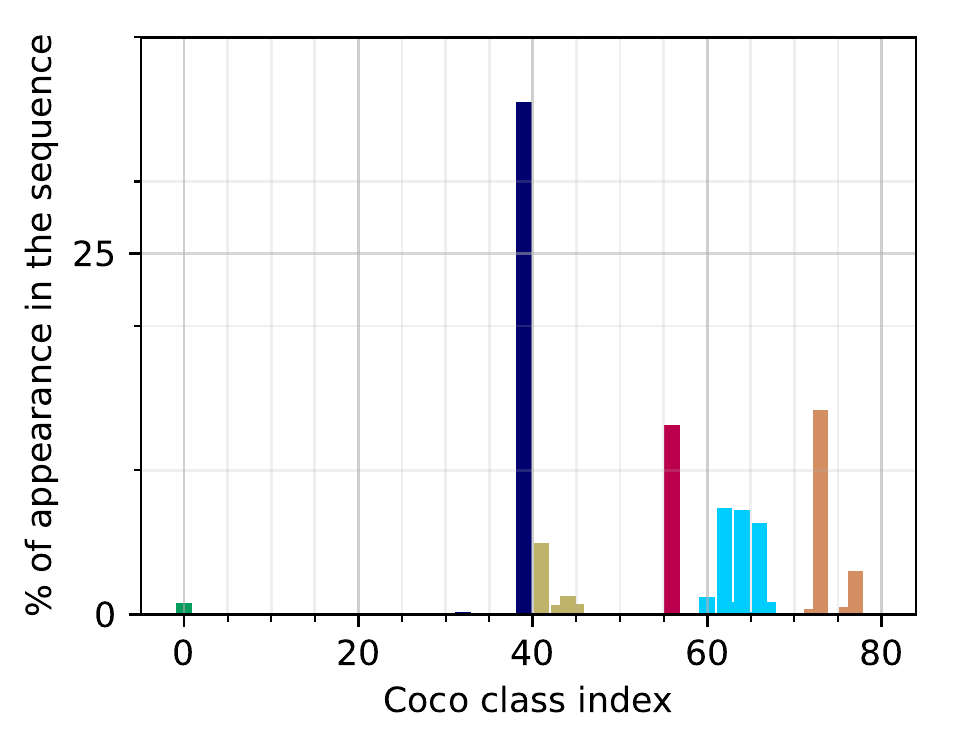}
\caption{Distribution of classes of detected objects during the full sequence with 10 robots. Bars of the same colors represent labels assigned to the same robot.}
\label{fig:class_dist}
\end{figure}

Fig.~\ref{fig:class_dist} shows that the classes observed in the sequence are
unbalanced, and generate higher load for specific robots with our
solution. For example, bottles (index 39) are more frequent in our
sequence. However, this distribution is meaningful and interpretable: specific
relations between objects are common in the environment. For example,
keyboards are usually close to screens, and chairs to tables. This may make
our method fragile against semantic perceptual aliasing, but recent robust
optimization techniques help in that regard \cite{lajoie_modeling_2019}.
Moreover, certain areas may be associated with specific objects, as
acknowledged by research on place categorization~\cite{wu_visual_2009}. We
believe the use of knowledge concerning the distribution of objects in the
environment can improve load balancing as future work.  We would also like to
underline that the use of constellations can, in theory, allow for relative
pose estimation between two frames. This step is currently performed through
visual features, which turns out to be the most bandwidth-consuming module of
DSLAM~\cite{cieslewski_data-efficient_2018}. Constellations have the potential
to greatly improve this step. However, in our experiments, the accuracy and
consistency of the object-to-point detector were not sufficient to produce
relative pose estimates of satisfying precision.

\section{CONCLUSIONS}
We present a new method to perform decentralized place recognition
using 3D constellations of objects. These constellations have the advantages
of being compact, meaningful, and inheriting robustness and invariance
associated with object detectors. Our solution maintains
performance when used with a decentralized mechanism in a sequence of the TUM RGB-D SLAM
dataset and matches state-of-the-art results in terms of required
bandwidth.

\bibliographystyle{IEEEtran}
\bibliography{IEEEabrv,bib} 

\end{document}